\documentclass[review]{elsarticle}

\usepackage{amssymb}
\usepackage{amsmath}
\usepackage{amsthm}
\usepackage{gensymb}
\usepackage{colortbl}
\usepackage{caption}
\usepackage{subcaption}
\usepackage{multirow}
\usepackage{acronym}
\usepackage{comment}
\DeclareMathOperator{\E}{\mathbb{E}}

\usepackage{bm}
\usepackage{graphicx}

\journal{Pattern Recognition}

\begin{document}

\begin{frontmatter}

\title{Graph Embedding with Data Uncertainty}

\author[mymainaddress]{Firas Laakom\corref{mycorrespondingauthor}}
\cortext[mycorrespondingauthor]{Corresponding author}
\ead{firas.laakom@tuni.fi}
\author[mysecaddress]{Jenni Raitoharju}
\ead{jenni.raitoharju@environment.fi}
\author[mymainaddress]{Nikolaos Passalis}
\ead{nikolaos.passalis@tuni.fi}
\author[mysecondaryaddress]{Alexandros Iosifidis}
\ead{ai@eng.au.dk}
\author[mymainaddress]{Moncef Gabbouj}
\ead{moncef.gabbouj@tuni.fi}

\address[mymainaddress]{Faculty of Information Technology and Communication Sciences, Tampere University, Tampere, Finland}
\address[mysecaddress]{Programme for Environmental Information, Finnish Environment Institute, Finland}
\address[mysecondaryaddress]{Department of Engineering, Aarhus University, Aarhus, Denmark}

\begin{abstract}
spectral-based subspace learning is a common data preprocessing step in many machine learning pipelines. The main aim is to learn a meaningful low dimensional embedding of the data. However, most subspace learning methods do not take into consideration possible measurement inaccuracies or artifacts that can lead to data with high uncertainty. Thus, learning directly from  raw data can be misleading and can negatively impact the accuracy. In this paper, we propose to model  artifacts in training data using probability distributions; each data point is represented by a Gaussian distribution centered at the original data point and having a variance modeling its uncertainty. We reformulate the Graph Embedding framework to make it suitable for learning from distributions and we study as special cases the Linear Discriminant Analysis and the Marginal Fisher Analysis techniques. Furthermore, we propose two schemes for modeling data uncertainty based on pair-wise distances in an unsupervised and a supervised contexts. \end{abstract}

\end{frontmatter}
\section{Introduction}
With the advancement of data collection processes, high dimensional data are available for applying machine learning approaches. However, the impracticability of working in high dimensional spaces due to the \textit{curse of dimensionality} and the realization that the data in many problems reside on manifolds with much lower dimensions than those of the original space, has led to the development of spectral-based subspace learning (SL) techniques. Spectral-based methods rely on the eigenanalysis of Scatter matrices. 
SL aims at determining a mapping of the original high-dimensional space into a lower-dimensional space preserving properties of interest in the input data. This mapping can be obtained using unsupervised methods, such as Principal Component Analysis (PCA) \cite{PCA1,park2009theoretical}, or supervised ones, such as Linear Discriminant Analysis (LDA) \cite{iosifidis2013rvda} and Marginal Fisher Analysis (MFA) \cite{4016549}.  Despite the different motivations of these spectral-based methods, a general formulation known as Graph Embedding was introduced in \cite{4016549} to unify them within a common framework. 

For low-dimensional data, where dimensionality reduction is not needed and classification algorithms can be applied directly,  many extensions modeling input data inaccuracies have recently been  proposed \cite{18,tzelepis2017linear}. In \cite{tzelepis2017linear}, data points are replaced by probability distributions modeling the artifacts and an SVM classifier was extended to operate on data distributions. However, for high dimensional data, where dimensionality reduction is needed, traditional methods, such as LDA and MFA do not take into consideration that the provided data can be exposed to measurement inaccuracies or artifacts. Thus, learning directly from data can lead to a biased or erroneous embedding of the high dimensional data \cite{gajamannage2019nonlinear,pan2009weighted,18,tzelepis2017linear}. Extensions of some SL methods taking into account the presence of outliers and noise in the data were proposed to account for this problem, such as the methods in \cite{saeidi2015uncertain,zheng2019l1} for LDA, and the method in \cite{vaswani2018robust} for PCA.

\begin{figure}
\centering
\includegraphics[width=0.49\linewidth]{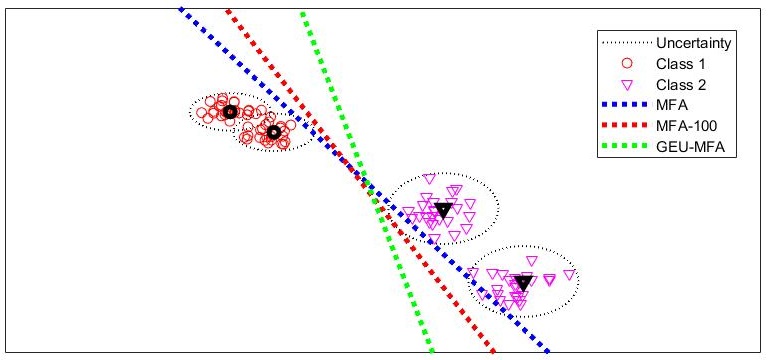}
\includegraphics[width=0.49\linewidth]{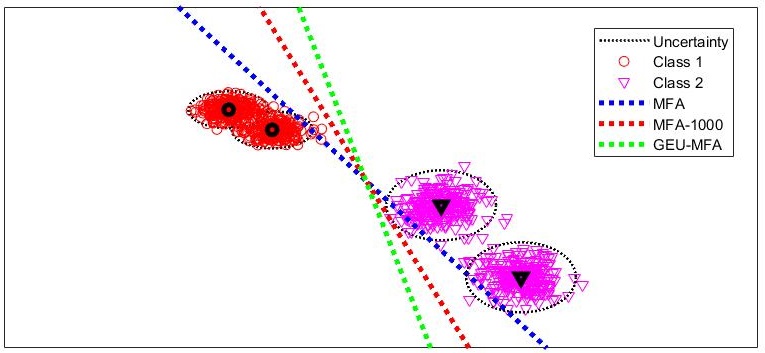}
\caption{The decision functions obtained by using MFA, GEU-MFA and MFA applied on augmented data by 100 samples, i.e., MFA-100 (left) and 1000 samples, i.e. MFA-1000 (right).}
\label{fig_1}
\end{figure}

In this paper, we propose a novel spectral-based subspace learning framework, called Graph Embedding with Data Uncertainty (GEU), in which input data uncertainties are taken into consideration. Instead of relying on the training data directly, we model each data point by a multivariate Gaussian distribution centered at the position of the original measurement and having a covariance matrix accounting for its uncertainty. To this end, we reformulate the Graph Embedding framework to operate on distributions at individual data point level allowing us to determine a mapping from the input data space into a lower-dimensional space via optimizing some properties of interest defined over these distributions. The outcome is a more robust data embedding scheme. As special cases  of the proposed framework formulations, we investigate extensions of LDA and MFA techniques within the proposed GEU framework. We refer to these as GEU-LDA and GEU-MFA, respectively. An example of the decision boundaries obtained by using the original MFA, MFA with augmented data, and GEU-MFA on 2-D synthetic data forming two classes is illustrated in Figure \ref{fig_1}. The incorporation of data uncertainty shifts the decision boundary of the original approach. We note that by using more augmented data the decision boundary of MFA shifts toward the  GEU-MFA. 

Furthermore, we theoretically show that under the proposed GEU framework, the rank of matrices involved in the optimization problem, i.e., the scatter matrices, increases compared to the original methods. As a result, methods formulated under the proposed framework lead to an increased number of projection directions. This is because the covariances employed to model the uncertainty at the  level of the individual data point introduce a regularization term to both scatter matrices. Thus, an indirect advantage of formulating traditional SL methods, such as LDA, under the proposed framework is that it allows for addressing the small sample size problem \cite{huang2002solving}, even for problems formed by two classes.

Although the focus in this paper in on LDA and MFA, the proposed GEU framework operating on generic graph structures can directly be used to obtain robust solutions for other SL methods formulated under the Graph Embedding framework. The contributions of the paper are as follows:

\begin{itemize}
    \item We propose a novel spectral-based subspace learning framework which takes into consideration uncertainties in the input data.
    
    \item We reformulate the Graph Embedding framework to operate on distributions at individual data points. In this way, we provide a generic approach for accounting for data uncertainties in a multitude of SL methods expressed under the Graph Embedding framework. 
    
    \item We study as special cases of the proposed framework GEU-LDA and GEU-MFA, and we theoretically show that considering uncertainty leads to an increased number of projection directions.
    
    \item We propose two schemes to model uncertainty of each sample based on pair-wise distances of data points in the original space.
    
\end{itemize}
The remainder of the paper is organized as follows. Section \ref{sec:related} provides a brief review of the related work. Section  \ref{sec:Proposed1} describes in detail the proposed GEU framework. Section \ref{sec:expirements} provides the conducted experimental analysis, and Section \ref{sec:conc} concludes our work.

\section{Related work} \label{sec:related}
\subsection{Graph Embedding}

Graph Embedding \cite{4016549,mygdalis2016graph,iosifidis2016multi} is a general framework encapsulating several SL methods as special cases. Data points are modeled as vertices of two graph structures, namely an intrinsic graph expressing data relationships to be emphasized and a penalty graph expressing data relationships  to be suppressed. Using such intrinsic and penalty graphs, the optimization problems of SL methods, such as LDA, PCA, and MFA, can be formulated. 

Given a set of data points and their corresponding class labels $\{ (\textbf{x}_i,c_i ) \}_{i=1}^N $, where $\textbf{x}_i \in \mathbb{R} ^{D}$ for $i= 1,...,N$, the goal in Graph Embedding is to determine a mapping which maps $\textbf{x}_i$ to a lower dimensional representation $\textbf{y}_i \in \mathbb{R} ^{d} $, $d < D$. This is achieved by forming a weighted (intrinsic) graph $G = \{\textbf{X}, \textbf{W}\}$, where $\textbf{X} = [\textbf{x}_1, . . . , \textbf{x}_N ]$ is the vertex set and $\textbf{W} \in \mathbb{R} ^{N \times N}$ the graph weight matrix whose elements encode the pair-wise relationships between the graph vertices $\textbf{x}_i$. Furthermore, a penalty graph $G^p = \{\textbf{X}, \textbf{W}^p\}$ can be defined on the same graph vertices, whose weight matrix $\textbf{W}^p \in \mathbb{R} ^{N \times N}$ expresses pair-wise relationships to be penalized.

The graph preserving criterion is formulated as follows:  
\begin{equation} \label{equ41}
\textbf{y}^*= {\underset{\textbf{y}^T\textbf{B}\textbf{y}=m}{\arg\min }\:\: \sum_{i\neq j} ( y_i - y_j )^2\textbf{W}_{ij} },
\end{equation}
where $\textbf{y}=[y_1,...,y_N]^T$, $y_i \in \mathbb{R}$ is a 1-D mapping of $\textbf{x}_i$, $m$ is a constant and \textbf{B} can be defined as a constraint matrix, e.g., $\textbf{B} = \textbf{I}$ to enforce orthogonality constraints, or as a scatter matrix based on the Laplacian of the penalty graph. For a linear data mapping, i.e., $\textbf{y} = \textbf{X}^T \textbf{v}$, where $\textbf{v} \in \mathbb{R}^D$ is a unitary projection vector mapping $\textbf{x}_i \in \mathbb{R}^D$ to $y_i \in \mathbb{R}$, Eq. (\ref{equ41}) can be rewritten as follows: 
\begin{equation}
\textbf{v}^* = {\underset{\textbf{v}^T \textbf{X}\textbf{B} \textbf{X}^T \textbf{v}=m}{\arg\min}\:\: \textbf{v}^T\textbf{X}\textbf{L}\textbf{X}^T\textbf{v} }, \label{Eq:CriterionGElinear}
\end{equation}
where  $\textbf{L} = \textbf{D} - \textbf{W}$ is the Laplacian matrix with \textbf{D} being the diagonal degree matrix having elements $\textbf{D}_{ii} = \sum_{j \neq i} \textbf{W}_{ij}$, and $\textbf{B} = \textbf{X} \textbf{L}^p \textbf{X}^T = \textbf{X} (\textbf{D}^p - \textbf{W}^p) \textbf{X}^T$. In this case, the solution of the optimization problem in Eq. (\ref{Eq:CriterionGElinear}) is given by solving the generalized eigenvalue decomposition problem
\begin{equation}
\left(\textbf{X}\textbf{L}\textbf{X}^T \right)\textbf{v} = \lambda \left(\textbf{X} \textbf{L}^p \textbf{X}^T\right) \textbf{v} \label{Eq:EigGE}
\end{equation}
and keeping the eigenvector corresponding to the smallest (positive) eigenvalue. To obtain more than one projection direction, the corresponding projection matrix $\textbf{V} \in \mathbb{R}^{D \times d}$ is formed by the eigenvectors corresponding to the $d$ smallest eigenvalues.

Specific selections of $\textbf{W}$ and $\textbf{W}^p$ lead to different subspace learning methods. For LDA, the within-class scatter and the between-class scatter matrices are given by
\begin{equation}
    \textbf{S}_w = \textbf{X}\left(\textbf{I} - \sum_{c=1}^{C} \frac{1}{N_{c}} \textbf{e}^{c} \textbf{e}^{cT}  \right)  \textbf{X}^T,
\end{equation}
\begin{equation}
\textbf{S}_b = \textbf{X} \left( \sum_{c=1}^{C} N_{c} \Big( \frac{1}{N_{c}}\textbf{e}^{c} - \frac{1}{N}\textbf{e} \Big)\Big( \frac{1}{N_{c}}\textbf{e}^{c} - \frac{1}{N}\textbf{e} \Big)^T \right) \textbf{X}^T,  
\end{equation}
where $C$ is the number of classes, $N_c$ is the cardinality of class $c$, $\textbf{e}\in R^N$ is the vector with all elements equal to $1$, and $\textbf{e}^c \in R^N$ is a vector with the elements corresponding to data points of class $c$ equal to one and the rest equal to zero. Thus, LDA can be formulated in the Graph Embedding framework by using the graph weight matrices
\begin{eqnarray}
    \textbf{W}_{ij} &=&  \left\{
                \begin{array}{ll}
 \frac{1}{N_{c_i}},& \textrm{if} \hspace{0.1cm} c_i = c_j \hspace{0.1cm} \textrm{and} \hspace{0.1cm}  i \neq j \\

 0, & \textrm{otherwise} \\
  \end{array} \right. \\
    \textbf{W}^p_{ij} &=&  \left\{
                \begin{array}{ll}
 \frac{1}{N}- \frac{1}{N_ {c_i}},  & \textrm{if} \hspace{0.1cm} c_i = c_j  \hspace{0.1cm} \textrm{and} \hspace{0.1cm}  i \neq j  \\
 \frac{1}{N},  & \textrm{otherwise} \\
  \end{array}
              \right.
\end{eqnarray}
where $N_{c_i}$ is the cardinality of the class, which $\textbf{x}_i$ belongs to. MFA is formulated by using the graph weight matrices
\begin{eqnarray}
    \textbf{W}_{ij} &=&  \left\{
                \begin{array}{ll}
 1,& \textrm{if} \hspace{0.1cm} i \in N^+_{k1}(j)  \hspace{0.1cm} \textrm{or} \hspace{0.1cm}  j \in N^+_{k1}(i) \\

 0, & \textrm{otherwise} \\
  \end{array} \right. \\
    \textbf{W}^p_{ij} &=&  \left\{
                \begin{array}{ll}
 1,& \textrm{if} \hspace{0.1cm} (i,j) \in P_{k_2}(c_i)  \hspace{0.1cm} \textrm{or} \hspace{0.1cm} (i,j) \in P_{k_2}(c_j) \\

 0, & \textrm{otherwise} \\
  \end{array}
              \right.
\end{eqnarray}
where $N^+_{k1}(j)$ is the set of the $k_1$ nearest neighbors of the $\textbf{x}_j$ in the same class, and $ P_{k_2}(c)$ is the set the $k_2$ nearest pairs among the set $\{(i,j),\textbf{x}_i\in c,\textbf{x}_j \not\in c \} $. Here, we should note that several other methods which employ pair-wise similarity/distance measures, e.g. \cite{pan2009weighted,bouzas2015graph, passalis2017dimensionality,yang2019nonparametric,wang2019learning, ornek2019nonlinear,mygdalis2016graph,  aytekin2017learning}, can be formulated using the Graph Embedding framework.

\subsection{Learning with uncertainty}
Research in uncertainty has gained a lot of attention lately in many branches of science  \cite{aliali,lourencco2017uncertainty}, since data can be subject to measurement inaccuracies and artifacts. Taking this into consideration in the data modeling and learning process is critical for building robust models. Exploiting uncertainty in machine learning has been studied from many different viewpoints. Methods dealing with uncertainty can be grouped into two different categories: sample-wise uncertainty modeling and feature-wise uncertainty modeling. 

In sample-wise uncertainty, the noise is modeled at the sample level. The main assumption in such methods is that few training data points are outliers and thus they need to be suppressed or partially suppressed to not affect the solution of the subsequent processing steps. Various robust extensions of SL methods have been proposed to reduce the sensitivity of a classifier to outliers \cite{saeidi2015uncertain,zheng2019l1,vaswani2018robust,wen2018robust,li2020robust,yue2019robust,xu2010robust,gajamannage2019nonlinear}.  In \cite{wen2018robust} and \cite{li2020robust} for example, robust extensions of LDA were proposed by reducing the sensitivity of the model to outliers. 

In feature-wise uncertainty, the noise is modeled at the data dimension level. The main assumption in such methods is that certain data dimensions are corrupted by noise. This type of noise modeling was employed to extend SVM in \cite{tzelepis2017linear}. For SL, feature-wise uncertainty is used in \cite{saeidi2015uncertain}, where a robust extension of LDA is proposed. Instead of using point estimates of speech data, a probabilistic description based on Gaussian distributions at the individual data point level are used as inputs to LDA. In our work, we use a similar uncertainty modeling. However, we note two key differences: i) Our work is based on the Graph Embedding framework formulation of SL and, thus, it is not restricted to LDA. ii) We propose two schemes to model the uncertainty of each sample based on pair-wise distances of data points in the original space. Thus, our approach of modeling uncertainty is not restricted to speech data and can be applied to any data, even when an explicit noise propagation model is absent.

\section{Graph Embedding with Data Uncertainty} \label{sec:Proposed1}
Let us denote by $\{ \underline{y}_i\}_{i=1}^N$ a set of the random Gaussian variables expressing the low-dimensional representations of the input data $\textbf{x}_i, \:i=1,\dots,N$. We express the graph preserving criterion using $\underline{y}_i$ as follows:
\begin{equation} \label{guass1}
\underline{\textbf{y}}^*= {\underset{\E(\underline{\textbf{y}}^T\textbf{B}\underline{\textbf{y}})=m}{\arg\min }\: \sum_{i\neq j} \E\Big( ( \underline{y}_i - \underline{y}_j)^2 \Big)\textbf{W}_{ij}  }, 
\end{equation}
where $\E(\cdot)$ denotes the expectation operator. For a Gaussian uncertainty, i.e.,  $\underline{y}_i \sim \mathcal{N}({\mu_i},\,{\sigma_i^2})$, the pair-wise distances $\underline{z}_{ij}$ between $\underline{y}_i$ and $\underline{y}_j$ are also random variables following a Gaussian distribution
\begin{equation}
\underline{z}_{ij} = \underline{y}_i- \underline{y}_j \sim \mathcal{N}({\mu_i - \mu_j},\,{\sigma_i^2+ \sigma_j^2}). 
\end{equation}
Thus, the expectation term in Eq. (\ref{guass1}) can then be  rewritten as follows:
\begin{eqnarray}\label{Eq:expectedDiff}
\E(( {\underline{y}}_{i} - {\underline{y}}_j)^2)  &=&  \E(\underline{z}_{ij}^2) = \E(\underline{z}_{ij})^2 + Var(\underline{z}_{ij})  \nonumber \\
&=& ( \mu_i - \mu_j)^2 +  (\sigma_i^2+ \sigma_j^2).
\end{eqnarray}
By substituting Eq. (\ref{Eq:expectedDiff}) to Eq. (\ref{guass1}), we get
\begin{eqnarray}
\underline{\textbf{y}}^* &=&   {\underset{\E(\underline{\textbf{y}}^T\textbf{B}\underline{\textbf{y}})=m}{\arg\min }\: \sum_{i\neq j} \E\Big( ( \underline{y}_i -\underline{y}_j)^2 \Big)\textbf{W}_{ij} } \nonumber \\
 &=&  {\underset{\E(\underline{\textbf{y}}^T\textbf{B}\underline{\textbf{y}})=m}{\arg\min }\: \sum_{i\neq j}   \Big( ( {\mu}_{i} - {\mu}_j )^2 + (\sigma_i^2+ \sigma_j^2) \Big) \textbf{W}_{ij} } 
\end{eqnarray}
The first term of the summation is equivalent to the original Graph Embedding and depends on $\E(\textbf{y})$, i.e., the expectation of \textbf{y}:
\begin{equation} \label{gauss3}
 \sum_{i\neq j}   ( {\mu}_{i} - {\mu}_j )^2  \textbf{W}_{ij} = 2\E(\textbf{y})^T \textbf{L} \E(\textbf{y}).
\end{equation}

By defining $\mbox{\boldmath$\sigma$} = \left[\sqrt{\sigma_1^2}, ... , \sqrt{\sigma_i^2} ,\sqrt{\sigma_n^2}\right]$, the second term in the summation can be expressed as follows: 
\begin{equation} \label{gauss2}
 \sum_{i\neq j} (\sigma_i^2+ \sigma_j^2)   \textbf{W}_{ij}=  2 \mbox{\boldmath$\sigma$}^T \textbf{D} \mbox{\boldmath$\sigma$}.
\end{equation}
Thus, using Eq. (\ref{gauss3}) and  Eq. (\ref{gauss2}), our new graph preserving criterion is given as follows:  
\begin{equation}
\underline{\textbf{y}}^* = {\underset{\E(\underline{\textbf{y}}^T\textbf{B}\underline{\textbf{y}})=m}{\arg\min }\:  \E(\textbf{y})^T \textbf{L} \E(\textbf{y}) + \mbox{\boldmath$\sigma$}^T \textbf{D} \mbox{\boldmath$\sigma$} }.  \label{gauss_final}
\end{equation}

For a linear data mapping $\underline{\textbf{y}}= \underline{\textbf{X}}^T \textbf{v}$ and modeling each data point in the input space using a Gaussian distribution, i.e., $\underline{\textbf{x}}_i \sim \mathcal{N}(\mbox{\boldmath$\mu$}^x_i,\mbox{\boldmath$\Sigma$}^x_i)$, ${\underline{y}}_i = \textbf{v}^T \underline{\textbf{x}}_i$ corresponds to a linear projection of a Gaussian, which is a Gaussian distribution $y_i \sim  \mathcal{N}(\mu^y_i,(\sigma^y_i)^2)$ with $\mbox{$\mu$}^y_i = \textbf{v}^T \mbox{\boldmath$\mu$}^x_i$ and $(\sigma^y_i)^2 = \textbf{v}^T \mbox{\boldmath$\Sigma$}^x_i  \textbf{v}$. Thus, the second term in Eq. (\ref{gauss_final}) can be written as follows:
\begin{equation}
\mbox{\boldmath$\sigma$}^T \textbf{D} \mbox{\boldmath$\sigma$} = \textbf{v}^T \left(\sum_{i} \textbf{D}_{ii}\mbox{\boldmath$\Sigma$}^x_i \right)  \textbf{v}. \label{Eq:sigmaDsigma}
\end{equation}

The equality in Eq. (\ref{Eq:sigmaDsigma}) follows from: $\mbox{\boldmath$\sigma$}^T \textbf{D}  \mbox{\boldmath$\sigma$}= \sum_{i}\sigma_i \sum_{j} ( \textbf{D}_{ij} \sigma_j)$. Since $\textbf{D}$ is diagonal, $\sum_{j} ( \textbf{D}_{ij} \sigma_j)= \textbf{D}_{ii}\sigma_i$.  Thus, $\mbox{\boldmath$\sigma$}^T \textbf{D} \mbox{\boldmath$\sigma$} = \sum_{i} \sigma_i^2  \textbf{D}_{ii}$. In addition, $\sigma_i^2 =\textbf{v}^T \mbox{\boldmath$\Sigma$}^x_i  \textbf{v} $, thus $\mbox{\boldmath$\sigma$}^T \textbf{D} \mbox{\boldmath$\sigma$}  = \textbf{v}^T \left(\sum_{i} \textbf{D}_{ii}\mbox{\boldmath$\Sigma$}^x_i \right)  \textbf{v}$.

Based on the above, the final form of Eq. (\ref{gauss_final}) is
\begin{equation}
\textbf{v}^*=  {\underset{\E(\textbf{v}^T\underline{\textbf{X}} \textbf{B} \underline{\textbf{X}}^T \textbf{v})=m}{\arg\min }\: \textbf{v}^T \left( \E(\textbf{X})^T \textbf{L}  \E(\textbf{X}) + \sum_{i} \textbf{D}_{ii}\mbox{\boldmath$\Sigma$}^x_i \right) \textbf{v} }. \label{Eq:finalCriterion}
\end{equation}
Following a derivation similar to the above, we note that a similar graph preserving criterion can be formulated with the constraint:
\begin{equation}
  \textbf{B}  = \left( \E(\textbf{X})^T \textbf{L}^p  \E(\textbf{X}) + \sum_{i} \textbf{D}^p_{ii}\mbox{\boldmath$\Sigma$}^x_i \right).
\end{equation}

The solution of the optimization problem in Eq. (\ref{Eq:finalCriterion}) is given by solving the following eigenvalue decomposition problem
\begin{equation}
\left( \E(\textbf{X})^T \textbf{L}  \E(\textbf{X}) + \sum_{i} \textbf{D}_{ii}\mbox{\boldmath$\Sigma$}^x_i \right) \textbf{v} = \lambda \textbf{B} \textbf{v} \label{Eq:FinalEig}
\end{equation}
and keeping the eigenvector corresponding to the smallest (positive) eigenvalue. To obtain more than one projection directions, the corresponding projection matrix $\textbf{V} \in \mathbb{R}^{D \times d}$ is formed by the eigenvectors corresponding to the $d$ smallest eigenvalues.

From Eq. (\ref{Eq:finalCriterion}), we can observe that when uncertainty is not used, i.e., by having   $\mbox{\boldmath$\Sigma$}^x_i$ equal to zero, the Gaussian distributions $\underline{\textbf{x}}_i$ become equivalent to Dirac function. Hence, in that case, Eq. (\ref{Eq:finalCriterion}) becomes equivalent to Eq. (\ref{Eq:CriterionGElinear}) and the solution of the proposed approach is equivalent to that of the original Graph Embedding framework. It should be noted that, as explained above, the projected data $\underline{y_i}^*$ obtained for each data point $\textbf{x}_i$ is also a random variable characterised by the mean $\E(y_i) = \textbf{v}^T \mbox{\boldmath$\mu$}^x_i$ and  variance $\sigma^y_i = \textbf{v}^T \mbox{\boldmath$\mu$}^x_i  \textbf{v}$. One can use this additional information for the projected data or only employ the first order approximation, i.e., the mean $\E(y_i)$, as the final projection of the original sample $\textbf{x}_i$. In this paper, we use the latter in the classification step. 

\subsection{Exploiting data uncertainty as a form of regularization}
By observing the eigenanalysis problem in Eq. (\ref{Eq:EigGE}), we can see that the number of projection directions which can be defined by the Graph Embedding framework depends on the underlying structure of the intrinsic and penalty graphs. That is, the maximal number of projection directions is upper bounded by the smallest rank of matrices $\textbf{X}\textbf{L}\textbf{X}^T$ and $\textbf{X} \textbf{L}^p \textbf{X}^T$. For example, when expressing LDA through Graph Embedding the maximal number of projection directions is equal to the rank of $\textbf{S}_b = \textbf{X} \textbf{L}^p \textbf{X}^T$, i.e., $\min(D,C-1)$, where $C$ is the number of classes. This restricts the number of meaningful projection directions that can be defined, leading to the extreme case of only one projection direction for binary problems. In order to solve the generalized eigenanalysis problem in Eg. (\ref{Eq:EigGE}), a regularized version $\tilde{\textbf{S}}_b = \textbf{X} \textbf{L}^p \textbf{X}^T + \epsilon \textbf{I}$ with $\epsilon > 0$ is used, because the original $\textbf{S}_b$ is singular. However, this regularization procedure simply shifts the eigen-spectrum of $\textbf{S}_b$ from $\lambda_i$ to $\tilde{\lambda}_i = \lambda_i + \epsilon \ge 0, \:i=1,\dots,D$) and has no data-driven intuition.

From Eq. (\ref{Eq:FinalEig}) we can see that both matrices involved in the generalized eigenanalysis problem of the proposed approach are strictly positive definite. That is, the additional terms $\sum_{i} \textbf{D}_{ii}\mbox{\boldmath$\Sigma$}^x_i$ and $\sum_{i} \textbf{D}^p_{ii}\mbox{\boldmath$\Sigma$}^x_i$ introduced to the scatter matrices defined over the intrinsic and penalty graphs act as regularization terms leading to full-rank matrices. This is due to that the Gaussian distribution covariance matrix, $\mbox{\boldmath$\Sigma$}^x_i$, is a strictly positive-definite matrix. Hence, the introduction of the proposed approach to model uncertainty at the individual data point level results in an intuitive regularization procedure, increasing the number of projection directions. This allows avoiding the small sample size problem of LDA \cite{huang2002solving} and provides more projection directions, even for binary problems.

\subsection{Uncertainty estimation} 
In the proposed GEU framework, we encode the uncertainty of each individual data point by a Gaussian distribution centered at the position of the data point and having a variance which needs to be appropriately determined to reflect the properties of the problem at hand. However, data is commonly available without such uncertainty information. We propose two schemes for defining such a variance estimate based on pair-wise distance between data points in the unsupervised and the supervised settings.

Each sample $\textbf{\underline{x}}_i$ is defined by its mean $\E(\underline{\textbf{x}_i})= \textbf{x}_i$ for both techniques and  its covariance $\bm{\Sigma_i}$ defined as follows:
\begin{equation}
\bm{\Sigma_i} = \sigma \: \textrm{diag}\Big( \textbf{x}_i -   \textbf{x}_{i^*} \Big)^2, 
\end{equation}
where $\sigma$ is a constant, diag$(\cdot)$ is the diagonal operator, and $\textbf{x}_{i^*}$ is the closest data point to $\textbf{x}_{i}$ in the admissible set. For the unsupervised case, the admissible set is composed of all the training data except $\textbf{x}_{i}$ and for the supervised case the admissible set is composed of all the training data except  $\textbf{x}_{i} $ and having the same class as $\textbf{x}_{i}$.

\begin{figure}[!t]
\centering
\includegraphics[width=\linewidth]{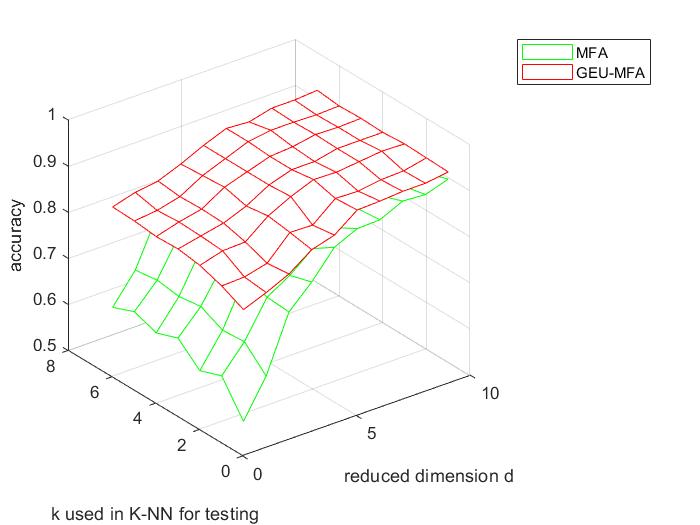}
\caption{Performance evaluation of MFA and GEU-MFA on Breast Cancer Wisconsin dataset for different combination of \textit{d}, the dimension of reduced space, and \textit{k} used in k-NN. }
\label{MFA1}
\end{figure}

\section{Experiments and analysis} \label{sec:expirements}
In this section, we study as special cases of the proposed framework the traditional subspace learning techniques LDA and MFA using our learning paradigm. For all testing scenarios, we rely on Nearest Neighbor for the classification. For the evaluation, we use three different datasets:
\begin{itemize}
    \item Breast Cancer Wisconsin dataset \cite{zhao1999subspace}:  It is a binary classification dataset composed of 569 samples with 32 features. An explicit uncertainty estimate is proposed in  \cite{tzelepis2017linear}. We use a random 5-fold split for the evaluation of different approaches. We keep the folds fixed for the different methods.
    \item Cifar2: We use two classes, ``cat'' and ``dog'', from the original Cifar10 \cite{krizhevsky2009learning}. We randomly sample 900 images per class for the training. For the testing, we use the original test set of  Cifar10 for both classes. To reduce the computational complexity, we first apply Bag of Visual Words (BoVW) using the SIFT descriptors to get a 400-dimensional representations of the original data.
    \item  Extended Yale B Face Database \cite{lee2005acquiring}: It contains 38 subjects and each subject provides 64 face images with different
    illumination conditions. Similar to \cite{wen2018robust}, we crop each image and convert it to a 32 by 32 gray image. Then, PCA is used to extract a 148 feature vector per sample. 
\end{itemize}
For all experiments, we cross-validate for the value of $\sigma$ from $\{0.001,0.1,0.2,0.4, 0.8,1,2\}$ and for the projection space dimension $d$ from  $\{1,2,4,8\}$. We denote the supervised and unsupervised variants of uncertainty estimation with S and U, respectively.

\subsection{MFA}
MFA is a SL technique which  characterizes  the  intraclass  compactness  in  the  intrinsic graph and the interclass separability in the penalty graph. It can be formulated using the Graph Embedding framework as explained in Section \ref{sec:related}. Thus, it can be extended using our framework to incorporate the data uncertainty using Eq. (18)-(20).

Figure \ref{MFA1} illustrates the performance of the original MFA and its uncertainty extension, i.e., GEU-MFA, for different combinations of reduced dimension $d$ and $k$ used in k-Nearest Neighbors (k-NN). We note that for small values of $k$ and $d$, GEU-MFA performs better than the original method. For the extreme case ($k=1, d=1$), MFA has 52.8\% accuracy compared to 77.1\% for GEU-MFA. For higher values of ($d$,$k$), the performance of both approaches increase and they tend to perform similarly.

\begin{figure}[!t]
\centering
\includegraphics[width=0.8\linewidth]{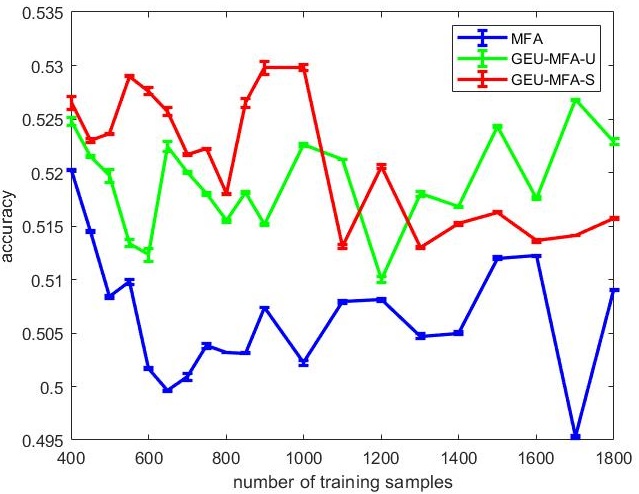}
\caption{Average accuracy  and variances of MFA, GEU-MFA-U, and GEU-MFA-S on Cifar2 for different training set sizes.}
\label{MFA2}
\end{figure}

\begin{figure}[!t]
\centering
\includegraphics[width=0.8\linewidth]{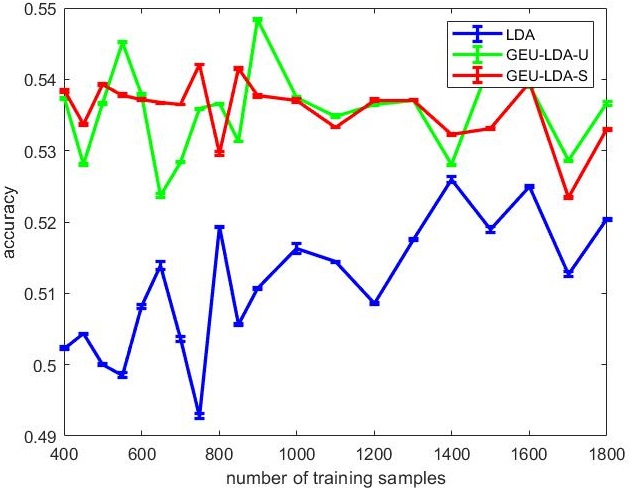}
\caption{Average accuracy  and variances of LDA, GEU-LDA-U, and GEU-LDA-S on Cifar2 for different training set sizes.}
\label{LDA2}
\end{figure}

\begin{table}[!t]
\setlength{\tabcolsep}{5pt}
\renewcommand{\arraystretch}{1}
	\caption{Classification accuracy of MFA \cite{4016549}, RMFA \cite{4016549}, GEU-MFA-U, and GEU-MFA-S in the different datasets.}
		\label{tab:mfatab}
\centering	
\begin{tabular}{l|r|c|c|c|c}
  &  noise  & MFA          &  RMFA  &  GEU-MFA-U    &  GEU-MFA-S \\
\hline
    & 0\% &0.858          &   0.851  &  0.866    &  \textbf{0.894} \\
Cancer  & 10\%  & 0.833          & 0.870 &  0.884   &  \textbf{0.890} \\
     & 20\%  & 0.806          &  0.825  &  0.835  & \textbf{0.849} \\
\hline
    & 0\% & 0.505          &  0.511  & 0.512    &  \textbf{0.520} \\
Cifar2  & 10\%  & 0.500          &  0.507 &  0.511    & \textbf{0.513} \\
     & 20\%  & 0.504          &   0.503 &  \textbf{0.506}    & \textbf{0.506} \\
\hline

    & 0\% &0.910          &   0.913  &  \textbf{0.922}   &  0.921 \\
Yale B face  & 10\%  & 0.901         & 0.902 &  0.905   & \textbf{0.910} \\
     & 20\%  & 0.892          &  0.896 &  0.901  & \textbf{0.902}\\
\hline

\end{tabular}
\end{table}

In  Figure \ref{MFA2}, we show the performance of the variants of MFA as a function of the number of training samples on Cifar2. We note that incorporating uncertainty consistently yields a performance boost for both variants of uncertainty techniques compared to the original MFA. For smaller training data sizes, the supervised variant usually leads to slightly better results (less than 1\%) than the unsupervised variant. When a higher number of training data is available, the unsupervised technique usually achieves the best accuracy.

In Table \ref{tab:mfatab}, we show the robustness of MFA \cite{4016549}, RMFA \cite{4016549}, and our proposed approach with both variants of uncertainty estimation, i.e., GEU-MFA-U and  GEU-MFA-S, on the three datasets with different additional  noise levels. We repeat each experiment ten times and report the average accuracy achieved by each method.  We note that the proposed methods outperform the original MFA for all noise levels. We also note that the accuracies of all the methods drop clearly when the noise level is higher. The supervised technique for estimating the uncertainty achieves the top performance except for Yale B Face dataset with no additional noise, where the best performance is achieved by GEU-MFA-U.

\subsection{LDA}

In  Figure \ref{LDA2}, we evaluate the performance of LDA, GEU-LDA-U and  GEU-LDA-S as a function of the number of training samples on Cifar2. We repeat each experiment ten times and report the mean and the variance of accuracies for all the training sizes.  Similar to MFA, incorporating uncertainty yields a performance boost for both variants of uncertainty techniques compared to the original LDA. We also note that for higher number of training samples, the performance gap decreases.  Both variants of uncertainty estimations achieve a similar performance for different training sizes.

We report the performance of LDA \cite{LDA}, regularized LDA \cite{4016549}, Robust Sparse Linear Discriminant  Analysis (RSLDA)  \cite{wen2018robust}, Uncertain Linear Discriminant  Analysis (ULDA) \cite{saeidi2015uncertain}, GEU-LDA-U, and  GEU-LDA-S on the three datasets for different noise levels in Table \ref{tab:final2}. We repeat each experiment ten times and report the average accuracy achieved by each approach. For the clean Cifar2 dataset, the best accuracy is achieved by GEU-LDA-U, while for the noisy Cifar2, GEU-LDA-S achieves the best results. The regularized LDA yields  the best accuracy for Cancer and Yale B (noise=10\%) datasets. However, for the other two variants of Yale B dataset, the highest accuracy is achieved by GEU-LDA-U. Compared to the original LDA, the LDA variants obtained via the proposed framework are more robust to the presence of noise and yield  higher accuracies.

\begin{table}[!t]
\setlength{\tabcolsep}{2.3pt}
\renewcommand{\arraystretch}{1}
	\caption{Classification accuracy of LDA \cite{LDA}, RLDA \cite{4016549}, RSLDA\cite{wen2018robust}, ULDA \cite{saeidi2015uncertain}, GEU-LDA-U, and GEU-LDA-S in the different datasets.}
		\label{tab:final2}
\centering	
\begin{tabular}{l|r|c|c|c|c|c|c}
 & noise & LDA  &  RLDA & RSLDA& ULDA  &  GEU-LDA-U    &  GEU-LDA-S \\
\hline
    & 0\% &0.523         &  0.541 & 0.511  & 0.505    &  \textbf{0.544} & 0.535 \\
Cifar2  & 10\%  & 0.497   & 0.538     & 0.516  & 0.501    &0.542 &\textbf{0.547}  \\
     & 20\%  & 0.523 & 0.545   &  0.510    & 0.498 &  0.541   &  \textbf{0.546} \\
\hline
    & 0\% & 0.932  &  \textbf{0.958}     & 0.882 &  0.528  & 0.951 & 0.950  \\
Cancer  & 10\%  &  0.896  &      \textbf{0.919}& 0.858 &0.541&  0.917 & 0.918 \\
     & 20\%  &0.895   &   \textbf{0.909}   &0.829  & 0.505    &  0.904 &  0.901  \\
\hline
    & 0\% & 0.856 &  0.869 & 0.851   &0.871 &\textbf{0.872} & 0.871  \\
Yale B  & 10\%  &  0.849    &   \textbf{0.864}& 0.827  & 0.859 &0.863 & 0.862\\
     & 20\%  &  0.838        & 0.853  & 0.839    & 0.852  &\textbf{0.856}& 0.855\\
\hline

\end{tabular}
\end{table}

\section{Conclusion} \label{sec:conc}
In this work, we introduced a novel spectral-based dimensionality reduction  framework called Graph Embedding with Data Uncertainty (GEU) that reformulates the Graph Embedding to consider input data uncertainties and artifacts.  We model the uncertainty around each data point by a multivariate Gaussian distribution centered around the original sample and a covariance matrix characterizing the uncertainty of the corresponding sample along each feature dimension. Two techniques to generate the distribution of each data point were proposed based on the pair-wise distances between samples.  Uncertainty introduces a regularization term that expands the rank of the scatter matrices and increases the number of available projection directions compared to the original subspace learning methods.  We studied as special cases of the proposed framework the traditional subspace learning techniques LDA and MFA. The proposed framework was extensively evaluated over three datasets and it led to performance improvement compared to the original methods as well competing methods that consider uncertainty.

\bibliography{mybibfile}

\end{document}